\documentclass{article}

\PassOptionsToPackage{numbers, compress}{natbib}

\usepackage[utf8]{inputenc} 
\usepackage[T1]{fontenc}    
\usepackage{hyperref}       
\usepackage{url}            
\usepackage{booktabs}       
\usepackage{amsfonts}       
\usepackage{nicefrac}       
\usepackage{amssymb}
\usepackage{amsfonts}
\usepackage{amsmath}
\usepackage{xcolor}         
\usepackage{theorem}
\usepackage{hyperref}
\usepackage{algorithm,algorithmicx,algpseudocode}
 \usepackage{enumitem}
 \usepackage{subfigure}
 \usepackage{bbm}
\usepackage[pdftex]{graphicx}

\usepackage[final]{neurips_2023}

\newcommand{\bE}{\ensuremath{\mathbb{E}}}
\newcommand{\bP}{\ensuremath{\mathbb{P}}}

\newcommand{\x}{\ensuremath{\mathbf{x}}}

\usepackage{amsmath,amsfonts,bm}

\def\1{\bm{1}}

\def\eps{{\epsilon}}

\DeclareMathAlphabet{\mathsfit}{\encodingdefault}{\sfdefault}{m}{sl}
\SetMathAlphabet{\mathsfit}{bold}{\encodingdefault}{\sfdefault}{bx}{n}

\newcommand{\probP}{\text{I\kern-0.15em P}}
\newcommand{\E}{\mathbb{E}}

\title{Accelerated Sampling of Rare Events using a Neural Network Bias Potential}

\author{%
  Xinru Hua \\
  Department of Computer Science\\
  Stanford University\\
  \texttt{huaxinru@stanford.edu} \\
  \And
  Rasool Ahmad \\
  Department of Mechanical Engineering \\
  Stanford University \\
  \texttt{rasool@stanford.edu} \\
  \AND
  Jose Blanchet \\
  Department of Management Science \\
  and Engineering \\
  Stanford University \\
  \texttt{jose.blanchet@stanford.edu} \\
  \And
  Wei Cai \\
  Department of Mechanical Engineering \\
  Stanford University \\
  \texttt{caiwei@stanford.edu} \\
}

\begin{document}

\maketitle

\begin{abstract}
In the field of computational physics and material science, the efficient sampling of rare events occurring at atomic scale is crucial. It aids in understanding mechanisms behind a wide range of important phenomena, including protein folding, conformal changes, chemical reactions and materials diffusion and deformation. Traditional simulation methods, such as Molecular Dynamics and Monte Carlo, often prove inefficient in capturing the timescale of these rare events by brute force. In this paper, we introduce a practical approach by combining the idea of importance sampling with deep neural networks (DNNs) that enhance the sampling of these rare events. In particular, we approximate the variance-free bias potential function with DNNs which is trained to maximize the probability of rare event transition under the importance potential function. This method is easily scalable to high-dimensional problems and provides robust statistical guarantees on the accuracy of the estimated probability of rare event transition. Furthermore, our algorithm can actively generate and learn from any successful samples, which is a novel improvement over existing methods. Using a 2D system as a test bed, we provide comparisons between results obtained from different training strategies, traditional Monte Carlo sampling and numerically solved optimal bias potential function under different temperatures. Our numerical results demonstrate the efficacy of the DNN-based importance sampling of rare events.

%
\end{abstract}

\section{Introduction}
Contemporary machine learning models suffer a substantial degradation in performance when confronted with long-tail events that are not represented well in collected data~\cite{FU2022290,10105457,NEURIPS2020_1e14bfe2}. Several approaches, including the detection of out-of-distribution samples~\cite{NEURIPS2019_1e795968} and data resampling techniques~\cite{chawla2002smote,buda2018systematic}, can address this issue. In this paper, we propose a method that is specifically designed to efficiently sample these long-tail events, also referred to as rare events. Our method aims to efficiently collect rare events in simulation and increase their representation in datasets.

Specifically, in the domains of materials science and bio-chemistry, there is a pressing need to sample rare events that are associated with specific physical phenomena and estimate their probabilities. This is critical in advancing our understanding of various materials properties ranging from mechanical composites~\cite{harada,doi:10.1021/acsapm.2c00503,doi:10.1080/08927020601052906} to transport proteins~\cite{doi:10.1126}, which are essential to aerospace and pharmaceutical industries. The characterization of rare event transitions between two stationary metastable states has been a focus of a big body of research in the past~\cite{vanden2006towards,vanden2010transition,gillilan1992shadowing}, and continues to stimulate the present research undertakings. In the same vein, we seek to study rare event transitions occurring in atomic systems with probabilities as low as $10^{-6}$. The molecules follow the Langevin dynamics governed by a potential energy function~\cite{schlick2010molecular} which contains multiple minima corresponding to metastable states. During the dynamics, the molecular system spends most of the time performing random thermal motion near a metastable state before rarely escaping to the neighboring metastable state. Our goal is to sample such rare event transitions of the molecular system between two metastable states of the potential energy function. The traditional Monte Carlo sampling method is prohibitively expensive to capture such rare events and suffers from the curse of dimensionality in higher dimensions. Another line of research focuses on determining the committor function $-$  the probability of making the rare event transition from a given molecular configuration $-$ by solving a partial differential equation numerically with finite element method~\cite{reddy2019introduction, alberty1999remarks}. However, such numerical determination of the committor function becomes exponentially difficult with the dimension of the problem. Thus, it is important to develop novel approaches to efficiently sample successful transitions and estimate the probability of rare events.

\begin{figure}[!ht]
\centering
\includegraphics[width=\textwidth]{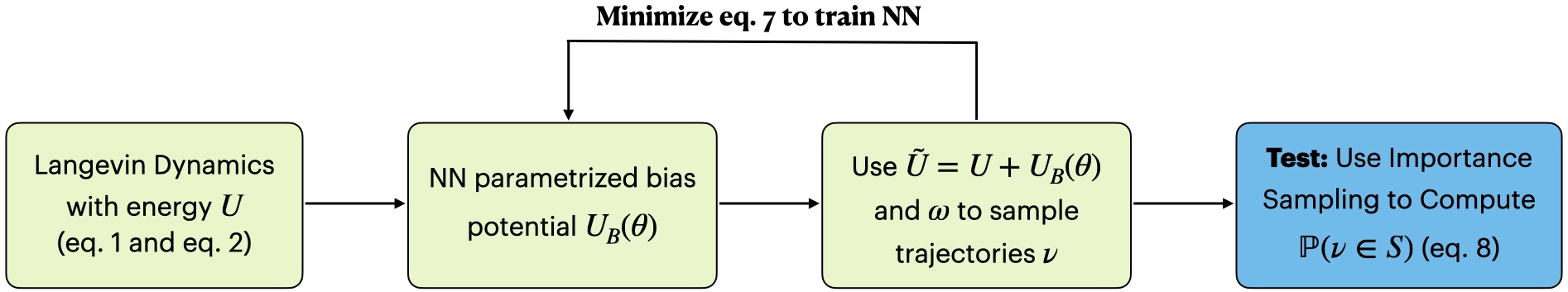}
\vspace{-10pt}
\caption{Full details of training are described in Algorithm~\ref{alg:bias} in Sec.~\ref{sec:algorithm}. Our method adopts a Reinforcement Learning approach: it samples trajectories and subsequently learns from them.}
\label{fig:flow_chart}
\end{figure}

In this work, we present an approach that leverages deep neural networks (DNNs) to learn from and sample distributions of rare events in molecular dynamics. A holistic view of our method is presented in Fig.~\ref{fig:flow_chart}. The method of applying bias potential to enhance sampling rate is first introduced in~\cite{de2005adaptive,cai2002importance} and both works present promising results on 2D. The limitations of both methods lie in the construction of a trial importance function and the challenges with multiple degrees of freedom in higher dimensions. This work \cite{hartmann2017variational} first formulate the estimation of rare event probabilities as an optimization problem and employed importance sampling techniques to recover the unbiased probabilities. Our method adopts a similar optimization framework and provides a practical framework to train, test, and evaluate the sample quality. One novelty of our method is its ability to learn from past successful transitions, so that we can add humans in the loop to sketch possible transition paths, use collected transition paths in real-world experiments~\cite{karplus,korlepara} or we can remove partially the energy barrier first and then obtain some successful paths. The optimal DNN-based bias potential need to both minimize the KL divergence between the distribution of transition paths under the biased dynamics and unbiased dynamics and maximize the probability of rare events. 

Our learning algorithm comprises three key steps: 1) The bias potential function is approximated by a DNN and the samples are generated by running the biased Langevin dynamics; 2) The DNN then learns to refine this sampled distribution and aligns it more closely to the unbiased distribution of rare event transitions; 3) After generating feasible rare events, we employ the importance sampling method to compute the actual unbiased probabilities of these rare events. Our paper makes several significant contributions:
\begin{enumerate}[leftmargin=6mm]
\item [\textbf{C1}] We introduce an efficient algorithm for training a DNN to approximate the variance-free importance/bias potential function. This trained DNN significantly improves the efficiency of our rare event sampling in molecular dynamics. Our method can effectively scale to higher-dimensional energy functions since it does not require data pre-collection or solving any equations.

\item [\textbf{C2}] Our algorithm effectively learn from previously successfully sampled rare events which gives our algorithm the freedom and power to reuse and learn from any distribution of trajectories.

\item [\textbf{C3}] 
Since we are sampling the molecule's trajectories, which exist in a space of more than 1000 dimensions, ensuring the statistical robustness of our estimator presents a significant challenge. To address this, we statistically assess the reliability of our estimator. This statistical measure enables us to compare the efficiency and accuracy of our estimator with other competing estimators in a fair and rigorous manner. 
\end{enumerate}
\section{Related Work}
\textbf{Optimal Control and Importance Sampling:} Several papers have delved into computing an optimal bias potential and utilizing importance sampling to estimate probability of rare events. Among them, a number of works focus on theoretical results on the optimal control and on transition paths~\cite{doi:10.1137/21M1437883,hartmann2017variational,hartmann2012efficient}, and the others provide numerical results~\cite{chak2023optimal,zhang2018reinforced,passerone2001action}. The main advantage of our work is the statistical guarantees of our estimator and the ability to learn from successful transitions. Additionally, our algorithm do not require data collection like in~\cite{zhang2018reinforced} and is scalable to higher dimension. In addition,~\cite{PhysRevE.66.046703} introduces the method to compute the bias potential to greatly enhance the rate of transitions, so that they can obtain the probability with much higher efficiency. However, their approach of computing the ground truth bias potential is restricted to dynamics on a grid. We have expanded upon this paper, adapting it to free dynamics in both 2D and higher dimensions. 

\textbf{Learning Committor Function:} Another popular line of research involves using neural networks to approximate the committor function, which requires solving a partial differential equation with neural networks~\cite{khoo2019solving,li2019computing,yuan2023optimal,pmlr-v145-li22a}. One notable limitation of this method is the significant growth in the size of the datasets and computational costs as the dimension grows. Specifically, it mandates the preparation of a dataset prior to the training. In contrast, our method can actively generate samples and learn from the samples simultaneously.

\textbf{Traditional Sampling Method:} There have been various sampling method to increase the efficiency of sampling low-probability events in molecular dynamics~\cite{frenkel2023understanding}, for example, the army ants algorithm~\cite{nangia2004army} and adaptive importance sampling method~\cite{de2005adaptive}. One shortcoming mentioned in~\cite{de2005adaptive} is difficulty of finding a suitable trial function for importance sampling that is successful in higher dimension and involving multiple degrees of freedom. Many books~\cite{prigogine2009advances} and review papers give a comprehensive summary to all the methods.

\section{Molecular Rare Event}

We study a class of rare events associated with transition paths in molecular dynamics of chemical reaction networks. The dynamics follow overdamped Langevin dynamics~\cite{pastor1994techniques,schlick2010molecular}:
\begin{equation*}
    d\x=-\nabla U(\x)/(m\gamma)\cdot dt+\sqrt{2k_BT/(m\gamma)}\cdot dB(t).
\end{equation*} Here $\x\in\mathbb{R}^d$, $k_B=8.617\times 10^{-5}$ is  Boltzmann's constant, $T$ is the absolute temperature, $m$ is the mass of the particle, $\gamma$ is the damping ratio, and $B(t)$ is a Brownian motion. For further simplification, we define $\eps=2k_BT/(m\gamma)$. To introduce the discretized dynamics with timestep of $\Delta t$, we first let $\omega=(\delta_0,\delta_1,...,\delta_N)$ to be a sequence of i.i.d noises, where $\delta_i\sim \mathcal{N}(0,\Delta t), \Delta t>0$. As a result, the discretized dynamics follows the equation:
\begin{equation}
    \Delta \x_t=-\nabla U(\x_{t-1})/(m\gamma)\cdot \Delta t+\sqrt{\epsilon}\cdot \delta_t.
    \label{eq:discretized_langevin}
\end{equation}
In this work, we focus on the 2D domain, but our method can be extended to general higher dimensions. We let $m=1,\gamma=1$, and an energy function $U$ on position $(x,y)$, defined as 
\begin{equation}
    U(\x)=0.05y+\frac{1}{6}\left(4(1-x^2-y^2)^2+2(x^2-2)^2+[(x+y)^2-1]^2+[(x-y)^2-1]^2-2\right).\label{eq:dynamics}
\end{equation} Figure 1 illustrates the energy function that has two potential wells separated by a potential barrier. In this figure, A and B represent the two minima of the energy function~\footnote{In the scenario where energy minima are unknown, we can use linear search and gradient descent method to determine energy minima.}. Our objective is to calculate the probability of a particle starting near A and reaching a region close to B before a set deadline. When the temperature is low, escaping the energy well around A becomes highly challenging, resulting in a very low success rate.
\begin{figure}[!ht]
\centering
\includegraphics[width=0.35\textwidth]{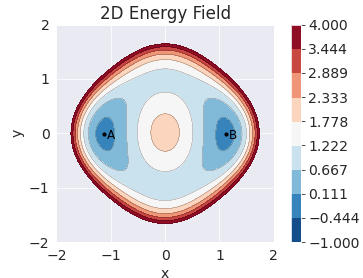}
\vspace{-5pt}
\caption{Plot of the 2D energy function in eq.~\ref{eq:dynamics}. A and B represent the two minima of the energy.}
\end{figure}

We aim to calculate the probability that a particle starts near A, escapes the energy basin, and moves to point B within a predetermined number of steps, denoted as $N\in\mathbb{N^+}$. We consider the particle to be close to B if its distance to B is within $\delta$. The particle forms an $N$-step trajectory $\nu = (\x_0, \x_1, ...,\x_N)$ as it follows the Langevin dynamics. In addition, we denote the step that the particle is closest to $B$ as $\tau_B$, that is
\begin{align*}
  \tau_B&=\min\{n:\|\x_n-B\|<\delta\}.
\end{align*}
The simulation stops when the particle reaches B or the deadline arrives. In this way, the rare event is defined as:
\begin{equation}\label{eq:rare_S}
    S = \{\nu:\tau_B\le\tau,\|\x_0-A\|<\delta\}.
\end{equation} 
From the original Langevin dynamics in eq.~\ref{eq:discretized_langevin}, we aim to learn a neural network with parameters $\theta$ to parameterize the bias potential function so the modified dynamics equation becomes
\begin{equation}
    \Delta \x(t)=-\nabla\left(U_B(\theta)(\x)+ U(\x)
    \right)\Delta t+\sqrt{\epsilon}\cdot \mathcal{N}(0,\Delta t).
\end{equation}\label{eq:biased_dynamics}The optimal modified dynamics significantly increases the probability of rare event $S$ and the successful trajectories are similar to the trajectories sampled with the original potential. Lastly, we employ importance sampling to recover the probability of the rare event with the original energy function $U$.

\textbf{Dimension of the problem:} Although the molecule's movement is restricted to a 2D plane, we aim to sample its trajectories. For each step in these trajectories, its position ($x$ and $y$) is recorded, resulting in a distribution with a dimensionality of $2N$, where $N$ is the number of steps in a trajectory. In our experiments, $N=500$, so the distribution is 1000-dimensional.

\textbf{Choice on $\tau$:} Generally, the time taken for a transition to occur is of the order $\mathcal{O}\left(\exp\left(\frac{1}{\eps}\right)\right)$, but the particle spends a substantial proportion of this time around A without crossing the energy barrier. Thus, if we want the particle to start from A and move to B without going back to A, we can select a deadline $\tau=\mathcal{O}\left(\frac{1}{\eps^r}\right)$ for some suitably chosen $r$, so that the particle has sufficient time to travel to B. More precisely, we choose $r$ such that $\tau>\E[\tau_B\;|\;\|\x_0-A\|<\delta,\tau_B<\tau_A]$.

\subsection{Variational Formulation of the Probability}
In this section, we explain how we write the rare event probability as a variational formula and obtain an optimization problem to train the neural network. After we sample $\omega$, we get our trajectory $\nu$, so we write trajectory as $\nu(\omega)$. Our goal is to express the probability $P(\nu(\omega)\in S)$ as the solution to an optimization problem. To do this, we introduce a function:
\begin{equation*}
F(\nu(\omega))=
\begin{cases}
  0, & \text{when}\ \nu\in S \\
  \infty, & \text{otherwise}
\end{cases}
\end{equation*}
so
\begin{equation*}
\exp(-F(\nu(\omega)))=
\begin{cases}
  1, & \text{when}\ \nu\in S \\
  0, & \text{otherwise}
\end{cases}
\end{equation*}

With this, we can calculate $P(\nu(\omega)\in S)=\mathbb{E}[\exp(-F(\nu(\omega)))]$. Suppose $Q$ is the new probability of trajectories that we sample $\nu$ with the modified potential function. Then, using the Jensen's inequality same as in~\cite{hartmann2017variational}, we can write the probability into a variational form:
\begin{align}
\log(\bP(\nu\in S))=-\inf\Bigl\{\mathbb{E}_Q \left[\log\left(\frac{dQ(\nu)}{dP(\nu)}\right)\right]+\bE_Q (F(\nu))\Bigl\}. \label{eq:optimization_raw}
\end{align}

We need to ensure that $dQ(x)=0$ if and only if $dP(x)=0$, and it is satisfied as they are both Gaussians. Here, $dP(\nu)$ denotes the probability distribution of trajectories controlled by the original energy function, and $dQ(\nu)$ represents the distribution with the modified energy function. Then,~\cite[Thm. 1]{hartmann2017variational} proves that the optimal solution to~\eqref{eq:optimization_raw} leads to a variation-free estimator for $\bP(\nu\in S)$.

\subsection{Smooth Indicator Function}\label{sec:smooth_ind}

From the definition of $F$, to minimize~\eqref{eq:optimization_raw}, we aim for every sampled $\nu$ to belong to $S$, so that $\E_Q(F)=0.$ 
To make the optimization approachable in practice, we replace $F$ by a smooth function: $F_{smooth}(\nu)=s\cdot \tanh\left(\|\nu(N)-B\|^2-(r+0.02)^2\right)$, where $\nu(N)=\x_N$, and $s$ and $r$ are both parameters. This function reaches its minimum at point B, increases smoothly as $x$ moves away from B, and remains constant when $|x-B|>r+0.02$. In our training scheme, $r$ decreases from 1.0 to 0.05 as we train the neural network, so trajectories need to move closer and closer to B.

\subsection{Likelihood Ratio Computation} We sample molecular trajectories by sampling $\omega$ and use the discretized dynamics model $\Delta\x_t=-\nabla \Tilde{U}(\x_{t-1})\cdot \Delta t+\sqrt{\eps}\cdot \delta_t$ with the modified potential function $\Tilde{U}=U+U_B(\theta)$ parameterized by a neural network $\theta$. At each step of the dynamics, we sample the random noise $\delta_t$ from a Gaussian distribution. In this way, the probability to model $\delta_i$ is the probability density functions (PDF) of the normal distribution, and we can write the likelihood of $\omega$ with respect to the Lebesgue measure as $p(\omega)$, which is a product of $N$ Gaussian PDFs. Since the randomness of $\nu$ is fully described by that of $\omega$, the likelihood of the trajectory $\nu$ equals to that of the corresponding $\omega$. 

In eq.~\ref{eq:optimization_raw}$, P, Q$ represent the distribution of molecular trajectories under two energy functions: $U,\Tilde{U}$. To compute $\frac{dQ}{dP}$, we first need to compute the corresponding $\omega_P$ and then compute the likelihood ratio $\frac{dQ}{dP}=\frac{p(\omega)}{p(\omega_P)}.$ For a trajectory $\nu=(\x_0,\x_1,...,\x_N)$, we first compute the Gaussian noises at each step as if it is generated by the original potential function $U$:
 \[\delta_t' = \frac{1}{\sqrt{\eps}}(\x_t-\x_{t-1}+\nabla U(\x_{t-1})\Delta t)
\]
Then, the likelihood ratio of the trajectory is 
\begin{align}
\log\left(\frac{dQ(\nu)}{dP(\nu)}\right)&= \log\left(\frac{p(\omega)}{p(\omega_P)}\right)=\log\left(\frac{p(\delta_N)p(\delta_{N-1})...}{p(\delta_N')p(\delta_{N-1}')...}\right)=\frac{1}{2\eps\, dt}\sum_{t=1}^{N} \left(\delta_t'^2-\delta_t^2\right)
 \label{eq:ratio}
 \end{align}

In practice, due to the time deadline and the fact that we smooth the indicator function, the bias potential may not always ensure that every trajectory successfully reaches point B. The final optimization problem then becomes:
\begin{equation}\label{eq:objective}
    \inf_{\theta}\bE_Q\left[\log\left(\frac{dQ(\nu_{\theta}(\omega)}{dP(\nu_{\theta}(\omega))}\right)\mathbbm{1}(\nu\in S)+F_{\rm smooth}\left(\nu_{\theta}(\omega)\right)\right],
\end{equation}
where $Q$ is the probability distribution of $\nu$ that is driven by the modified energy function $U+U_B(\theta)$.


\section{Algorithm}\label{sec:algorithm}
Using the bias potential, we want to see the likelihood of rare events significantly increases, and the distribution of successful trajectories remains unchanged. In our experiments, we use the energy function~\ref{alg:bias} as the loss function and batch stochastic gradient descent method~\cite{ruder2016overview} to train our DNN. After many simulation runs, the bias potential $U_B$ allow us to sample from a distribution of trajectories $Q$ that minimizes the objective function~\eqref{eq:objective}. Algorithm~\ref{alg:bias} describes the training process. In line 12, we minimize two components in the loss function: the first part is the KL divergence between $Q$ and $P$, which tries to match the distribution of successful trajectories under the modified and the original energy, and the second part penalizes unsuccessful trajectories.
\begin{algorithm}[!ht]
    \caption{Train a DNN as a bias potential function to sample from $P\left(\nu|\nu\in S\right)$}
    \label{alg:bias}
\algrenewcommand\algorithmicrequire{\textbf{Input:}}
\algrenewcommand\algorithmicensure{\textbf{Output:}}
\algnewcommand{\LineComment}[1]{\State \(\triangleright\) #1}
\algorithmicrequire{ Two minima A and B, learning rate $\alpha$, starting position $x_0$, original potential $U$, threshold $\delta$, time step $\Delta t$, number of simulations $M$, number of timesteps $N$.}

\algorithmicensure{ A bias potential function $U_B$ parameterized by neural network $\theta:\mathbb{R}^2\rightarrow \mathbb{R}$}
\begin{algorithmic}[1]
\For{$i = 1,2,\ldots, M$}
\For{$t=1,\ldots,N$}
\State $x_{t}=x_{t-1}-\nabla (U+U_B(\theta))(x_{t-1})\Delta t+\sqrt{\epsilon}\mathcal{N}(0,\Delta t)$. \Comment{Parallelized over a batch of samples.}
\For{each unfinished trajectory in the batch}
\If{$\|x_t-B\|<\delta$}
\State Compute the KL loss in eq.~\ref{eq:objective}: $L_{\text{KL}}=\mathbb{E}_Q \left[\log\left(\frac{dQ(\nu)}{dP(\nu)}\right)\right]$.
\State Mark this trajectory as finished.
\EndIf
\EndFor
\EndFor
\State Compute the $F\_{\text{smooth}}$ in eq.~\ref{eq:objective} as $L_{\text{smooth}}$ and $L_{\text{total}}=L_{\text{KL}}+L_{\text{smooth}}$.
 
\State Run gradient descent: $\theta = \theta-\alpha \nabla_{
\theta
}L_{\text{total}}$.
\EndFor
\State \Return $v_{\theta}$
\end{algorithmic}
\end{algorithm}



After training the neural network, we employ importance sampling to compute the rare event's probability under the original distribution $P$. We run simulations using the biased dynamics as described in eq.~\ref{eq:biased_dynamics}. As in eq.~\ref{eq:objective}, we define $Q$ to only model the successful trajectories. After computing $\omega_p$, the probability of $\nu\in S$ under $P$ is given by:
\begin{align}
    \bP(\nu \in S)=\E_Q\left[\frac{dP(\nu)}{dQ(\nu)}\right]=\frac{1}{M}\sum_{i}\frac{p(\omega_{p,i})}{p(\omega_i)}\mathbbm{1}(\nu_i(\omega_i)\in S).
    \label{eq:imp_sampling}
\end{align}
Here $P$ represents the distribution of $\nu$ under the original potential and $Q$ is the distribution of $\nu$ under the modified potential. The density ratio $\frac{p(\omega_p)}{p(\omega)}$ can be computed analogously to eq.~\ref{eq:ratio}.

To achieve an efficient estimation, we want the density ratio of successful trajectories $W_i=\frac{dP(\nu)}{dQ(\nu)}$ to be at the same scale with every sample, so that the estimation is not described by a limited subset of samples. Therefore, we use a metric called Effective Sample Size (ESS)~\cite{MARTINO2017386} and coefficient of variance (CV) to measure the efficiency of importance sampling: 
\begin{align*}
    {\rm ESS} = \frac{(\sum_{i:\nu_i\in S} W_i)^2}{\sum_{i:\nu_i\in S} W_i^2},\quad
    CV = \frac{\sigma}{\mu},\text{where $\sigma$ is the sample standard deviation}.
\end{align*}
Additionally, we utilize the confidence interval and standard deviation of the estimation to measure the reliability of our estimators.

\subsection{Learn from Existing Successful Trajectories}\label{sec:learn_exist}
In the case of higher dimensions or more transition channels, it will be extremely difficult to successfully sample trajectories in all the transition channels in one simulation run. One great advantage of our algorithm is that it is adapted to learn a bias potential from a dataset of successful paths. With this functionality, we can obtain some successful paths with any method first and then train a possible bias potential to speed up further sampling: (1) experts in material science can sketch possible transition paths based on their understanding, (2) we can remove the energy barrier partially and then obtain some successful paths, (3) we can also obtain successful paths from mechanical experiments. Curating a balanced dataset of successful paths helps the neural network learn to generate all modes of trajectories and be less biased. It also greatly speeds up the process of training a neural network.

\begin{figure}[]
    \centering
\begin{minipage}{0.4\textwidth} 
    \includegraphics[width=\linewidth]{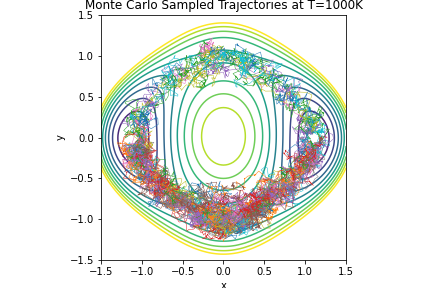}    
    \end{minipage}
\begin{minipage}{0.4\textwidth} 
        \includegraphics[width=\linewidth]{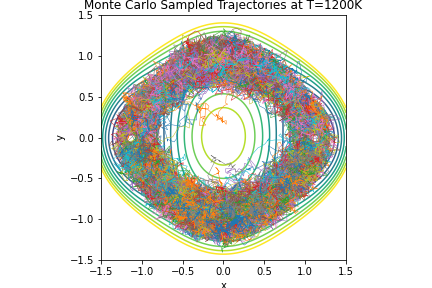}
    \end{minipage}
            \vspace{-5pt}
    \caption{Trajectories at temperature 1000K and 1200K obtained by traditional Monte Carlo sampling method. The molecules escape at a higher rate at the elevated temperature of 1200K. However, the two channels of transition are more precise and refined at the lower temperature. }
    \label{fig:traj}
\end{figure}

In the 2D problem, there are two channels that the molecule can transit through: going up and going down, like plotted in Fig.~\ref{fig:traj}. Since our energy function is asymmetric on $y$, it is more likely to escape the energy basin through the bottom channel. During training, sometimes the neural networks converge to local minima where they generate trajectories only going through one channel, so combining the two modes and learning from all trajectories is useful.

We minimize the KL divergence between the probability of the existing paths under the original potential ($p$) and the probability of these existing paths under the bias potential that is being learned ($q$). We continue using the existing paths and the neural network does not generate new paths. If we have a collection of successful paths: $\{\nu_i,i=1...n\}$, we can solve:
\begin{equation}
    \min_{\theta} \sum_i \log\left(\frac{q_{\theta}(\nu_i)}{p(\nu_i)}q_{\theta}(\nu_i)\right)
    \label{eq:exist_loss}
\end{equation}

\section{Results}
The neural network architecture consists of 4 hidden layers and $\tanh$ as the activation function (except the output layer). It maps a molecular position to the bias energy: $\mathbb{R}^2\rightarrow\mathbb{R}.$ We find that adding two control variables $\exp(\|x-A\|_2)$ and $\exp(\|x-B\|_2)$ allows the DNN to better utilize the information of the distance between $x$ and A or B. The project is implemented using the PyTorch library. The gradient of bias energy $\nabla U_B(\x)$ is computed by the PyTorch autograd module. All the experiments are performed on one AWS c5.4xlarge instance.

\begin{figure}[t]
    \centering
        \vspace{-5pt}
    \includegraphics[width=0.8\linewidth]{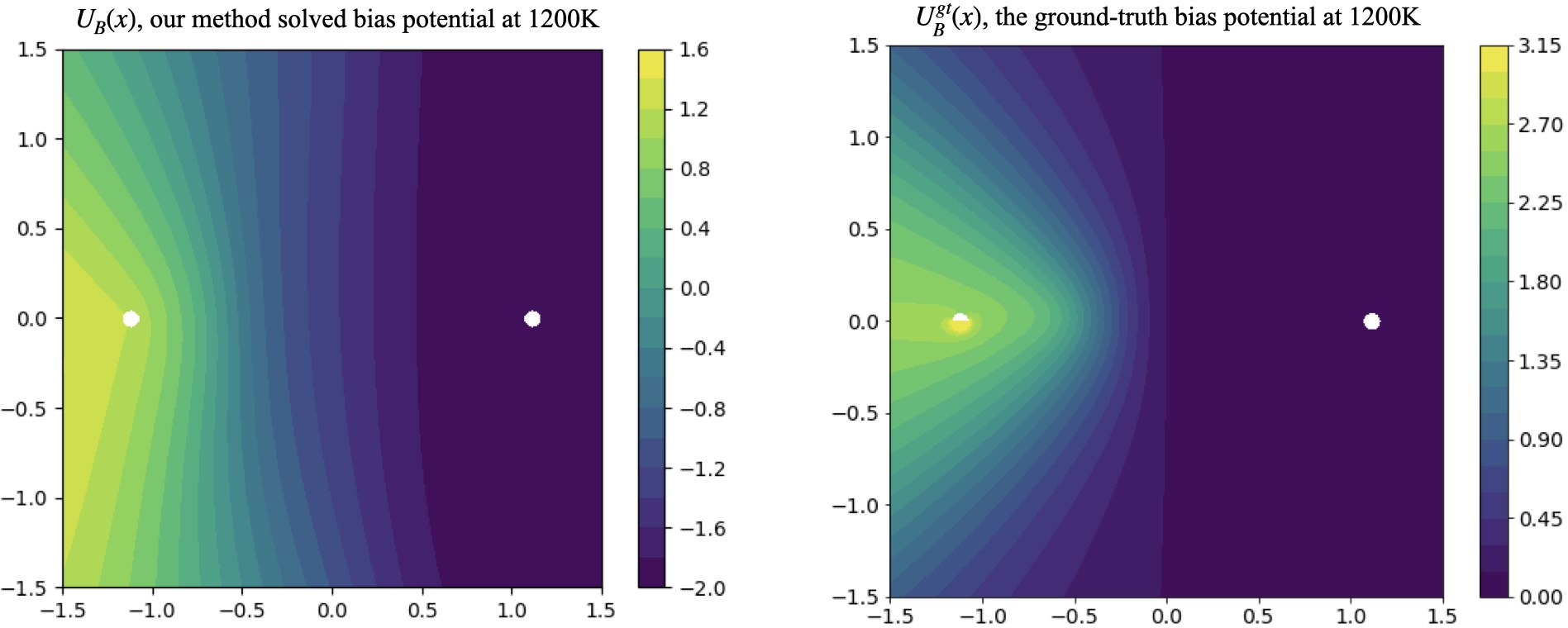}
    \vspace{-5pt}
    \caption{\textbf{Left:} Bias potential functions generated by our method in mode A. \textbf{Right:} Bias potential generated by the PDE's numerical solution. The shapes of two functions are very similar, although our optimization process does not involve solving the PDE.}
\label{fig:bias}
\end{figure}

\begin{figure}[t]
    \centering
        \vspace{-5pt}
    \includegraphics[width=\linewidth]{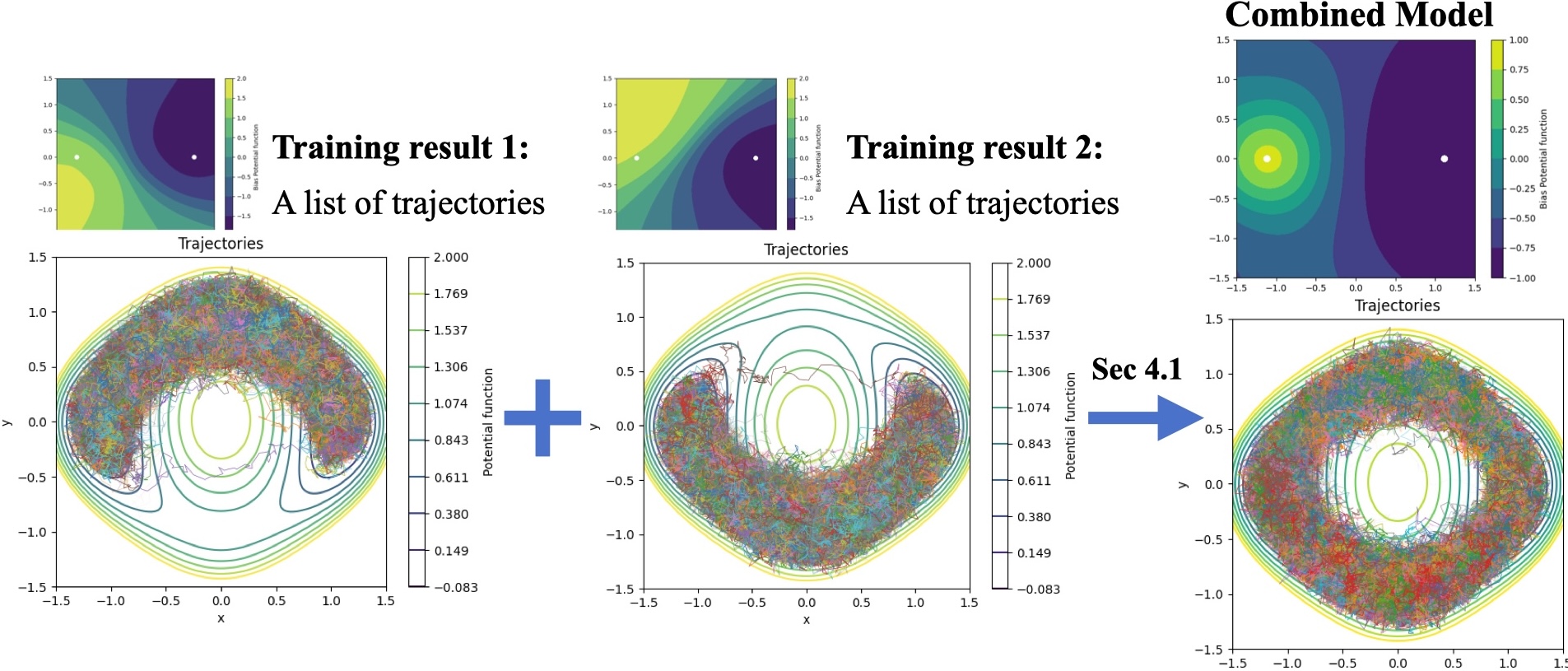}
    \vspace{-5pt}
    \caption{Mode B of our method at temperature 1200K. Training in mode A sometimes suffers from mode collapse as other generative methods. In this case, we only need the positions of the successful paths, and we minimize~\eqref{eq:exist_loss} to train a new bias potential that can sample paths that are similar to all the past training examples. The sampled trajectories are also similar to the Monte Carlo's in Fig.~\ref{fig:traj}. }
\label{fig:combine_training}
\end{figure}

Our method is a generative approach, so it does not require a standard train/val/test split. Every trajectory is sampled with random Gaussian noises. During the test, we freeze the DNN-based bias potential and use it in eq.~\eqref{eq:biased_dynamics} to sample trajectories. We train our method with a batch size of 512 and train 300 steps. The comparisons of the confidence interval, success rate, ESS, and time consumption under two temperature settings are demonstrated in Table~\ref{tab:1200} and Table~\ref{tab:1000}. We use a statistical significance at a P-value of 0.05 throughout the tests. We showcase two functionalities of our method: 

\paragraph{A: Exploration} We train a DNN from scratch following Alg.~\ref{alg:bias} with the loss function~\eqref{eq:objective}. 
We obtain the ground truth bias potential $U_B^{gt}(\x)=-2K_BT\log(q(\x))$ by numerically solving the committor function with finite element method (FEM)~\cite{reddy2019introduction} from the following partial differential equation (PDE)~\cite{alberty1999remarks,persson2004simple}:
\begin{equation} 
    \nabla\cdot\left(e^{-\beta U(x)}q(x)\right)=0, \text{in}\ D\ \backslash \ 
 (A\cup B), \quad q(x)\vert_{\partial A}=0,\quad q(x)\vert_{\partial B}=1.
\end{equation}\label{eq:PDE}
In higher dimensions, solving the committor function from this PDE becomes infeasible~\cite{khoo2019solving}. Thus, it is critical to develop other methods that does not require solving the PDE with FEM or other numerical methods in higher dimensions. A comparison between our method's bias potential and the ground truth bias potential is in Fig.~\ref{fig:bias}.
\paragraph{B: Combine} Given successfully sampled trajectories, we combine all the past knowledge into one DNN with the loss function in eq.~\ref{eq:exist_loss}. The trajectories can be from other DNNs, the Monte Carlo method, or human experts. We only need the positions of the trajectories, so that we can combine the knowledge into one bias potential. The pipeline and the results are shown in Fig.~\ref{fig:combine_training}.
\begin{table}[!ht]
  \caption{Comparison between our method A, B and Monte Carlo method under 1200K. We test our method with 5120 examples and the Monte Carlo method with $10^8$ samples. Even though our method does not directly minimize the variance of the estimator, the variance from our method A is smaller than Monte Carlo. ESS is not applied to Monte Carlo. All computational times recorded are measured in minutes. Combining training and testing, our method A achieves a 5.13x speedup over naive Monte Carlo and our method B achieves 2.5x speedup. \\ \textbf{Note:} Our training time is an upfront cost and does not grow as the number of samples grows. If only comparing the test time, our method offers a 44x speedup.}
  \label{sample-table}
  \centering
  \begin{tabular}{llllllll}
    \toprule
    Temperature 1200K    & Confidence Interval  &CV &Success Rate & ESS ratio& $t_{\text{train}}$ &$t_{\text{test}}$\\
    \midrule
Our method A &$4.037\pm 0.342\mathrm{e}{-6}$ &3.0933 &0.770 &0.095 &122 &16\\
Our method B &$3.232\pm 0.743\mathrm{e}{-5}$ &8.402 &0.396 &0.014 &14 &16 \\
Monte Carlo   &$4.410\pm 0.412\mathrm{e}{-6}$ &1505.846 &$4.410\mathrm{e}{-6}$ &- &- &708\\
    \bottomrule
  \end{tabular}
  \label{tab:1200}
\end{table}
\begin{table}[!ht]
\caption{Similar to Table~\ref{sample-table}, we compare the results between different methods. Our method A also has smaller variance and 4.4x speed up compared to naive Monte Carlo.}
\centering
\begin{tabular}{llllllll}
\toprule
Temperature 1000K     & Confidence Interval &CV &Success Rate & ESS ratio & $t_{\text{train}}$ &$t_{\text{test}}$\\
\midrule
Our method A &$3.433\pm0.285\mathrm{e}{-7}$ &3.032 &0.819& 0.098 &132 &15\\
Our method B &$4.993\pm0.759\mathrm{e}{-6}$ &8.093 &0.396 &0.015 &10 &16 \\
Monte Carlo   &$3.600\pm1.176\mathrm{e}{-7}$ &5270.463 &$3.600\mathrm{e}{-7}$&- &- &646 \\
\bottomrule
\end{tabular}
\label{tab:1000}
\end{table}

\subsection{Robustness of Estimator}
\begin{figure}[t]
\centering
\includegraphics[width=\linewidth]{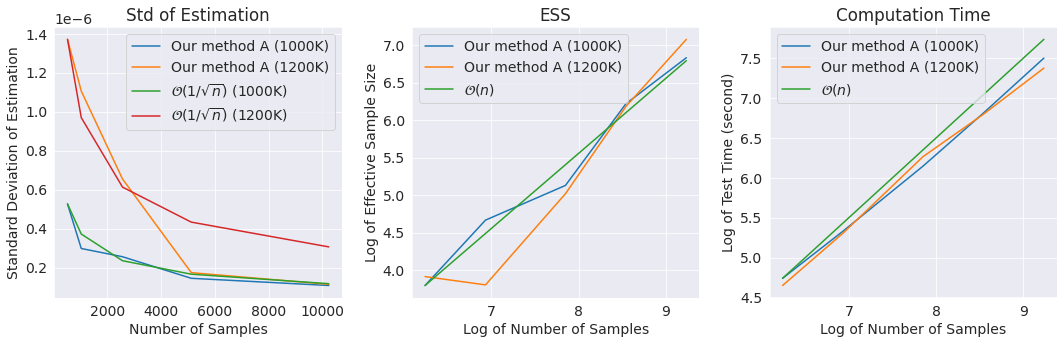}
\caption{\textbf{Left:} Our method achieves $\mathcal{O}(1/\sqrt{n})$ reduction in the standard deviation of population mean, the same rate as naive Monte Carlo. Here the numbers for 1000K are multiplied with 10 to be at the same scale as the numbers of 1200K. \textbf{Middle:} The ESS increases in the order of $\mathcal{O}(n)$, which means the effective sample size is proportional to the sample size, with the proportionality constant being stable. \textbf{Right:} Our method's computation time exhibits a growth rate of $\mathcal{O}(n)$.} 
\label{fig:scale}
\end{figure}

In this section, we plot the scaling behavior of our estimator, focusing on variance, ESS, and computational time as functions of the number of test samples in Fig.~\ref{fig:scale}. Our findings reveal that our estimator mirrors the trend observed in Monte Carlo methods. Specifically, as the number of samples escalates, there is a reduction in variance coupled with an enhancement in ESS. Notably, the testing time increases slower than linearly, due to the considerable overhead associated with initializing and loading the neural network bias potential. This underscores the robustness and reliability of our approach in comparison to traditional Monte Carlo.

\section{Conclusion}
In this work, we focus on challenges of efficiently sampling rare events in molecular dynamics, which underpins the technologically important properties of materials and molecules. We propose to utilize deep neural networks as a bias/importnace potential function. We successfully formulate the probability as an optimization problem which is used to train the neural network. Our proposed approach can be scaled  to high-dimensional cases and also provides robust statistical guarantees on the accuracy of the estimated probabilities. The ability of our algorithm to learn from successful samples makes our method versatile and marks an improvement over existing methodologies. We compare our method with traditional Monte Carlo sampling and numerical FEM method under different temperatures to measure the efficacy of our approach. Our estimator exhibits a smaller variance than the traditional Monte Carlo and achieves about 5x speedup comparing the training and test time combined, and more than 44x speedup if only comparing the test time. We also test the robustness and scalability of our bias potential. With more test samples increasing, the variance decreases in the order of $\mathcal{O}(1/\sqrt{n})$, and effective sample size and time increase in the order of $\mathcal{O}(n)$.

The immediate future direction is applying our method to higher dimensional models, similar to~\cite{voter1997hyperdynamics}. In higher dimensions, we anticipate our method to offer even greater speed advantages over the naive Monte Carlo approach, which suffers from the curse of dimensionality. Another intriguing avenue to explore is the potential for training neural networks using existing molecular dynamics datasets, as described in~\cite{frassek2021extended,korlepara}. One limitation of our method lies in the variance of likelihood ratio of importance sampling. We plan to test the stability and sensitivity of the importance sampling method by adding artificial noises to the bias potential and measuring how much the results change. We will also compare against another line of research~\cite{khoo2019solving,pmlr-v145-li22a}, which solves the committor function and tries to use it to enhance the sampling rate.
\newpage
\section{Acknowledgement}
We would like to express our gratitude to the anonymous reviewers for their insightful comments. Our appreciation extends to Professor Masha (Maria) K. Cameron for sharing her FEM code with us, and to Jiaxin (Margot) Yuan for valuable discussions. We are also thankful to Dr. Tianyi Shi for meticulously proofreading this manuscript and for meaningful discussions. The research is supported by the Air Force Office of Scientific Research under award number FA9550-20-1-0397. Additional support is gratefully acknowledged from NSF 1915967, 2118199, 2229012, 2312204. 

\bibliographystyle{ieeetr}
\bibliography{ref}
\end{document}